\documentclass[a4paper,10pt]{article}

\usepackage{amssymb}

\usepackage[figuresright]{rotating}

\usepackage[section]{algorithm}
\floatname{algorithm}{Algorithme}
\usepackage{algorithmicx}

\usepackage{amsmath}
\usepackage{ragged2e}

\newenvironment{keywords}{\textbf{Keywords : }}{}

\title{Regression Trees and Random forest based feature selection for malaria risk  exposure prediction.}
\author{ Bienvenue Kouway\`e$^{1,\; 2, \;   *}$}
\date{ 1- Universit\'e d'Abomey-Calavi, International Chair in Mathmatical Physic and Applications (ICMP:UNESCO-Chair), Abomey-Calavi, B\'enin\\ 
 2- Universit\'e Paris 1  Panth\'eon Sorbonne,  Laboratoire SAMM, Paris, France.\\
 * E-mail :   kouwaye2000@yahoo.fr
 }
\begin{document}

\maketitle
\justifying
\begin{abstract}
This paper deals with prediction of anopheles number, the main vector of malaria risk, using environmental and climate variables.
The variables selection is based on an automatic machine learning  method using regression trees, and random  forests combined with stratified 
two levels cross validation.
The minimum threshold of variables importance is accessed  using the quadratic distance of variables importance while 
the optimal subset of  selected variables is used to perform predictions. Finally the results revealed to be qualitatively better, at the 
selection, the prediction,
and the CPU time point of view than those obtained by GLM-Lasso method.
\end{abstract}


\section{Introduction}
\label{Introduction}
Generally,  studies about disease like  chikungunya, aids, and malaria provide data set  containing  a high number of variables and a small number of observations.
When it is important to perform  prediction on the risk of these diseases, the goal is to provide a consistent heuristic to select the probable candidate predictor and 
 perform prediction for the study and also where only  explanatory   data are  available. Generally, experts in medicine, epidemiology, genetic,
perform treatment on variables before analysis,  operations of selection,  and forecast. Based on their knowledge, they decide to 
transform some variables in classes, to fix interactions between some variables, etc.
In epidemiology context, the aim of this work is to provide an automatic algorithm for variables selection based on regression trees and random forest.  
This procedure must overcome the treatment done by experts, generate automatically a stable, and optimal subset of predictors,  and perform
prediction for an other area where the target variable is not available.
There are a lot of statistical modeling approach, such as linear model (LM), linear mixed model (LMM),
generalized linear (GLM), generalized linear mixed (GLMM) for selection or prediction. However, these models fail when $p>n$,
the number of variables ($p$) is more important  than the number of observations ($n$). Experts also 
 assume independence among explanatory variables. A lot  of methods of variables selection provide a subset for prediction but  not stable, 
not consistent or more demanding in computation time like \textit{Wrapper, embedded, filter, ranking}, and their  variant \cite{Guyon03anintroduction,DBLP:conf/esann/Bontempi05,Diaz-UriarteSelection}.
 In recent works, we proposed one method of variables selection based on combination of  Lasso and GLM 
through a double cross-validation  (LOLO-DCV) named GLM-Lasso \cite{Kouwayemldm, Kouwayejds}.
The present work combines the stratified double cross validation (LOLO-DCV), and regression trees or random  forest.
This implies two methods : LOLO-DCV combined with regression trees, and LOLO-DCV combined with random forest.
For malaria risk prediction, four strategies of variables
selection LDRT, LDCT, LDRF, and  LDCF are implemented. These strategies use some criteria such as : 
the mean, the quadratic risk, the absolute risk of the predictions, and 
the CPU time of algorithm computation.
Each strategy is applied on four groups of variables (original, original with village, recoded, recoded with village). 
Most of the algorithms implemented in our work are based on \cite{ Kouwayemldm, Kouwayejds, Friedman2015glmnet, goeman2010l1, zou2005regularization}.
We first provided a threshold of variable importance measurement for each strategy.
This threshold is very important because it puts out the importance or not of predictors. The second step is to make prediction through a double cross validation 
loop. The last step is to select predictors according to their frequency and make consistent prediction with the remain  predictors. 
The results are compared to those obtained by reference method.
The results obtained by such 
 procedure are clearly better improved compared
to those obtained by  GLM-Lasso \cite{Kouwayemldm, Kouwayejds} taken as the reference method. The improvement is about 
 all properties 
such as  the selection power, the selection accuracy, the sparsity of the best subset of variables, and the prediction. Moreover, 
the CPU time used to display
our program is smaller than the one 
required by the reference method and only few climate and environmental variables are the main 
factors associated to the malaria risk exposure with an improved accuracy.

\section{ Materials}
In this section, we briefly recall the description of the study area, the mosquito collection  
 and identification as well as the data,  
  and 
related variables. For more details, see \cite{cottrell2012malaria}.

\subsection{Study area}
The study was conducted in the district of Tori-Bossito
(Republic of Benin), from July 2007 to July 2009. Tori-Bossito is on the coastal plain of Southern Benin, 40 kilometers
north-east of Cotonou. This area has a subtropical climate and
during the study,  
 the rainy season lasted from May to October.
Average monthly temperatures varied between 27$^\circ$C and 31$^\circ$C.
The original equatorial forest has been cleared and the vegetation
is characterized by bushes with sparse trees, a few oil palm
plantations, 
 and farms. The study area contained nine villages
(Avam\'e centre, Gb\'edjougo, Houngo, Anavi\'e, Dohinoko, Gb\'etaga,
Tori Cada Centre, Z\'eb\`e, 
 and Zoungoudo). Tori Bossito was
recently classified as mesoendemic with a clinical malaria
incidence of about 1.5 episodes per child per year \cite{damien2010malaria}.
Pyrethroid-resistant malaria vectors are present \cite{Djenontin}.

\subsection{Mosquito collection and identification}
Entomological surveys based on human landing catches (HLC)
were performed in the nine villages every six weeks for two years
(July 2007 to July 2009). Mosquitoes were collected at four catch
houses in each village over three successive nights (four indoors and
four outdoors, i.e. a total of 216 nights every six weeks in the nine
villages). Five catch sites had to be changed in the course of the study
(2 in Gbedjougo, 1 in Avam\`e, 1 in Cada, 1 in Dohinoko) and a total
of 19 data collections were performed in the field from July 2007
to July 2009. In total, data from 41 catch sites are available.
Each collector caught of predictional 
 mosquitoes landing on the lower legs
and feet between 10 pm and 6 am. 
All 
 mosquitoes were held in
bags labeled with the time of collection. The following morning,
mosquitoes were identified on the basis of morphological criteria
\cite{Gillies1,Gillies2}. All An. gambiae complex and An. funestus mosquitoes were
stored in individual tube 
 with silica gel and preserved at 220$^\circ$C.
P. falciparum infection rates were then determined on the head and
thorax of individual anopheline specimens by CSP-ELISA \cite{Wirtz}.
\subsection{Environmental and behavioral data}
Rainfall was recorded twice a day with a pluviometer in each
village. In and around each catch site, the following information
was systematically collected: (1) type of soil (dry lateritic or humid
hydromorphic)—assessed using a soil map of the area (map IGN
Benin at 1/200 000 e , sheets NB-31-XIV and NB-31-XV, 1968)
that was georeferenced and input into a GIS; (2) presence of areas
where building constructions are ongoing with tools or holes
representing potential breeding habitats for anopheles; (3)
presence of abandoned objects (or ustensils) susceptible to be used
as oviposition sites for female mosquitoes; (4) a watercourse
nearby; (5) number of windows and doors; (6) type of roof (straw or
metal); (7) number of inhabitants; (8) ownership of a bed-net or (9)
insect repellent; And (10) normalized difference vegetation index
(NDVI) which was estimated for 100 meters around the catch site
with a SPOT 5 High Resolution (10 m colors) satellite image
(Image Spot5, CNES, 2003, distribution SpotImage S.A) with
assessment of the chlorophyll density of each pixel of the image.
Due to logistical problems, rainfall measurements are only
available after the second entomological survey. Consequently, we
excluded the first and second survey 
 (performed in July and
August 2007 respectively) from the statistical analyses.
\subsection{Variables}
The dependent variable was the number of
Anopheles collected in a house over the three nights of each
catch and the explanatory variables were the environmental
factors, i.e. the mean rainfall between two catches (classified
according to quartile), the number of rainy days in the ten days
before the catch (3 classes [0–1], [2–4], $>$4 days), the season
during which the catch was carried out (4 classes: end of the dry
season  from February to April; beginning of the rainy season from May to
July; end of the rainy season from August to October; beginning of the
dry season from November to January), the type of soil 100 meters
around the house (dry or humid), the presence of constructions
within 100 meters of the house (yes/no), the presence of
abandoned tools within 100 meters of the house (yes/no), the
presence of a watercourse within 500 meters of the house (yes/no),
NDVI 100 meters around the house (classified according to
quartile), the type of roof (straw or Sheet metal), the number of
windows (classified according to quartile), the ownership of bed
nets (yes/no), the use of insect repellent (yes/no), 
 and the number
of inhabitants in the house (classified according to quartile). These pre-treatments 
  based
on the knowledge of experts in entomology, and medicine operated 
on some  original variables  generate  a  second type of  variables called recoded variables. 
The  original and recoded variables are described in Table
 \ref{tab:Recoded variables}. 
Two  types of variables set are used :  the first set, the original variables with all variables obtained by interactions; 
The second set, the recoded variables with all variables obtained by interactions. For
knowing the effect of the village on the selection method and prediction, four groups of variables are considered : Group 1 (original variables), 
Group 2 (original variables with village as fixed effect),
Group 3 (recoded variables), and Group 4 (recoded variables with village as fixed effect)

\section{Methodology}
\label{Methodology}
\subsection{Variables importance}
For individual tree,  the variable importance  is defined used the out-of-bag sampling like :
\begin{equation}
 \label{var_impor_Tree}
VI^{T}(X^p)=  (err\widetilde{OOB}_t^p - errOOB_t)
\end{equation}
The naive importance measure in tree-based ensemble methods is to merely count the number 
 of times each variable is selected by all individual tree in the group of trees. There also exist 
 the "Gini importance" measurement used in random forest for classification. The more advanced variable
 importance measurement in random forest is the "permutation accuracy importance" defined as: 
 \begin{equation}
 \label{var_impor_RF}
VI^{RF}(X^p)= \frac{1}{ntree} \sum_{t=1}^{ntree} (err\widetilde{OOB}_t^p - errOOB_t)
\end{equation}
where $VI^{T}$ is variable importance of a tree, $VI^{RF}$ is variable importance of a random forest,
$X^p$ is the $p$-th variable, $OOB_t$ the out-of-bag of tree $t$, $\widetilde{OOB}_t^p$  the sample 
obtained by randomly permuting the value of $X^p$ in $OOB_t$ and $errOOB_t$ the out-of-bag error 
for the tree $t$, $ntree$ the number of regression trees in the random forest.

The Gini importance, and the permutation accuracy importance measures are employed as variable 
selection criteria in many recent study in various disciplines.
The effects induced by the differences in scale level of the predictors are more pronounced for
the  \textbf{randomForest} function, where variable selection in the individual tree is biased, than  the one with
 \textbf{cforest} function where the individual trees are unbiased \cite{Strobl2007,Archer:2008:ECR:1316079.1316183}. It has been  also 
  shown that if  \textbf{cforest} function is used with bootstrap sampling, the variables selection 
 frequencies of the categorical predictors still depend on their number of categories.
Variable importance has a sensitivity to the number of observations and the number of variables.
This sensitivity is reduced with increasing number of true variables.
Variable importance has also sensitivity to $mtry$ the minimum number of observations at a node  for splitting, and $ntree$ the maximal
number of trees in forest. It has been shown that for a fixed number of observations and variables, the effect of taking a larger value 
for $mtry$ is evident. Indeed, the magnitude of variable importance is more double starting from $mtry=14$ to $mtry = 100$, and it again 
increases with $mtry = 200$. The effect of $ntree$ is less visible  but taking $ntree=2000$ leads to better stability \cite{Genuer:2010:VSU:1857266.1857397}
A lot of strategies have been developed for variable selection. The recursive elimination of feature based on variable importance developed
by Avlarez de Andr\'e runs like this. They first compute random forest variable importance. Then, at each step they eliminate the $20\%$
of variables having the less importance and build a new forest with the remaining variables. They finally select the set of variables 
leading to the smallest $OOB$ error rate of a forest defined by 
\begin{equation}
\label{OOB_error}
 errOOB = \frac{1}{n}Card\{ i \in \{ 1, \ldots, n\} | y_i \neq \hat{y}_i\}
\end{equation}
where $\hat{y}_i$ is the most frequent label predict by trees $t$ for which 
$(x_i, y_i)$ is in the $OOB_t$ sample \cite{Diaz-UriarteSelection}. But this proposition 
of variables  elimination is arbitrary,  and the method does not depend 
on the data.

Robin Genuer proposed an other method of variable selection based on variable importance stratified in two steps \cite{Genuer:2010:VSU:1857266.1857397}.
The step 1 is a preliminary ranking which  consisting in  sorting the variables in decreasing order of Random forest
scores of importance, and  canceling the variables of small importance, $m$ is the number of remaining
at the second step, he selected the variables involved in the  model leading to the smallest
$OOB$ error and at the end constructed an ascending sequence of Random forest models by invoking, and testing the variables stepwise. 
The variables of the last model are selected.
But this method lacks of precision because in the step 1 this strategy is sensible when it exist irrelevant
variables, and at last step,  variables  invoking or testing can be sensitive to high correlation among variables.
The method to access variable importance proposed by Díaz-Uriarte \cite{Ramon-Diaz-uriarte} in scaled, unscaled, and Gini version is only available
when the dependent variable is a factor.
\subsection{Performance and accuracy in variables selection for strategies }
This part of the work is based on simulated data.  It is necessary to  show the power of each strategy to reduce
effectively the number of variables and select  the right variables  in the optimal subset for prediction. 
Let $\mathbf{V}^R$ and $\mathbf{V}^W$ the set of the real, and wrong 
variables respectively, $\mathbf{S}^R$, $\mathbf{S}^W$ the set of the real, and wrong selected
variables respectively. Let $  \mathbf{V}=  \mathbf{V}^R \cup \mathbf{V}^W$ and $  \mathbf{S}=  \mathbf{S}^R \cup \mathbf{S}^W$.

\subsubsection {Selection power} 
It is defined as the ratio of the number of  variables selected  on the number of total variables (real and wrong).
This quantity gives an idea of percentage of elimination of  variables.
The selection power is noted $SP$ and  defined as :
\begin{equation}
\label{Selction_power}
 SP = \frac{Card(\mathbf{S})}{Card(\mathbf{V})}
\end{equation}


\subsubsection {Selection accuracy} 
It is defined as the ratio of the number of real variables selected  on the number of total variables selected.
This quantity gives the accuracy selection of variables.
 It is  noted $SA$ and defined as :
 \begin{equation}
 \label{Selection_accuracy}
SA= \frac{Card(\mathbf{S}^R)}{Card(\mathbf{S})}
 \end{equation}

 
\subsection {Strategy parameters construction} 
The parameters used in the strategies for variables selection are, the minimum 
threshold of variables importance, the minimum number of observations at each node before 
splitting in trees,  and the maximum number of trees in forest building.

\subsubsection {Heuristic of variable importance measurement} 
The strategy of variable importance measurement proposed 
in this paper is based on a minimum threshold. For any model of regression trees and random forest, if any 
variable has importance greater than this threshold, it is 
considered as important variable in the model. 
One of the difficulties in this study is the decision of the minimum of importance of variable.
We have a lot of techniques to check this number. The strategy proposed by Genuer et al \cite{Genuer:2010:VSU:1857266.1857397} is 
sensible when it exist irrelevant variables. A classical alternative is to select  the threshold according to some elbow finding strategy on the 
variable importance mean curve. In this paper  we propose a new  strategy running like this : we run the full model using the whole data frame $n_r$ times 
(default $n_r=$100) with the default parameters. 
The matrix of variables important noted $M_{VI}$ is a $ n_r\times q$-matrix defined as :
\begin{center}
$M_{VI} = 
\begin{pmatrix}
VI_{11}&VI_{12}&\ldots VI_{1n_r}\\
VI_{21}&VI_{22}&\ldots VI_{2n_r}\\
\vdots &\vdots &  \vdots\\
VI_{q1}&VI_{q2}&\ldots VI_{qn_r}\\
\end{pmatrix}$
\end{center}
$VI_{ij}$ is the importance of i$^{th}$ variable at j$^{th}$ repetition, 
$1 \leq i \leq q$ and $1 \leq j \leq n_r$. 
 Let $$M_{VI}=\left(VI_{.1}, VI_{.2},\ldots,VI_{.n_r}\right)$$ 
 if $$VI_{.i}=\left( VI_{1i},VI_{2i},\ldots, VI_{qi} \right)^t$$  
 then   $$\sigma_i = min\{VI_{.i}, VI_{.i}\neq \theta, 1 \leq i \leq n_r \}$$ 
 and $$\sigma = (\sigma_1, \sigma_2,\ldots, \sigma_{n_r})$$ where
 $A^t$ is the transposed of the matrix  $A$, and $\theta$ is the null vector.
 The minimum threshold of variables importance noted $VI_{min}$ is  defined as :
\begin{equation}
\label{var_import_meas}
 VI_{min} =  min(\sigma)+sd(\sigma)
\end{equation}
\subsubsection {Parameters $mtry$ and  $ntree$ accessing} 
These parameters are accessed through a simple cross validation process.
The data set is divided into two parts : $E_A$ the learning set,  and $E_T$ the test set.
On $E_A$,  it has  been performed one kind of variable selection method varying one specific parameter.
For the regression  tree,  the parameter concerned $mtry$, the minimum number of observations 
that must exist at a node in order for a split to be attempted. For the forest, this parameter is  $ntree$, 
the maximal number of trees in the forest. These parameters will be noted $m$ for simplification of notations, 
 $1\leq m\leq n_{obs} $ where  $n_{obs} $ is the number 
of observations in $E_A$. For each value of $m$, the corresponding  regression  tree  or the random  forest 
provides a vector of 
importance $VI_{.\; m}$. Let $\widehat{VI}_{m}$ the vector of mean of the  vectors $VI_{.\; m}$.
The quadratic distance is  defined as :
 \begin{equation}
 \label{QD}
  d(\widehat{VI}_m,VI_{.\; m}) = \left[ \sum_{j=1}^{n_{var}} (\widehat{VI}_m -VI_{j\;m} )^2 \right]^{1/2}, \, 1 \leq m \leq n_{obs}
 \end{equation} 
 where $n_{var}$ is the number of variables
The parameter $m$ is determined by optimizing the quadratic distance of importance.
Let define :
\begin{eqnarray}
 \mathcal{H}&=&Arg\min_{m}{d(\widehat{VI}_m,VI_{.\; m})}
\end{eqnarray}
%
%
 If $Card(\mathcal{H}) =1$ then $\mathcal{H}=\{h_0\}$ and
 \begin{equation}
  mtry=ntree=h_0
 \end{equation}
If $Card(\mathcal{H}) \geq 2$ then 

\begin{equation}
 mtry= min \{\mathcal{H}\}\, \mbox{and}\,\,ntree= max \{\mathcal{H}\}
\end{equation}
A  regression tree with this value of $mtry$ or a Random forest (RF) with 
this value of $ntree$ will perform prediction on $E_T$. All this process
will be repeated until prediction is computed for all observations.

\subsubsection {Algorithm of variables  selection and prediction} 
This algorithm is similar to the one developed in
 a recent work \cite{Kouwayejds,Kouwayemldm}.

\begin{algorithm}{}
 \caption{LOLO-DCV-Tree-Forest}
 \begin{enumerate}
 \item Determination of $VI_{min}$
 \item 
The data are separated  in  $N$-folds
 \item A each step of the first level
 \begin{enumerate}
\item The folds are regrouped in two part : $E_A$ and $E_T$, 
$E_A$ : the learning set which 
contained the observations of $(N-1)$-folds,\\ $E_T$ : the test set, contained the observations of the last fold. 
\item Holding-out $E_T$
\item \label{CV1} The second level of cross-validation
\begin{enumerate}
\item    A full  cross validation is computed  on $E_A$ for determination of the first model construction  parameter $m$ ($mtry$ or 
$ntree$).
  \item  Tree or forest model $\mathcal{M}_{m}$ is computed on  $E_A$ using $m$
  \item The importance of variables ${VI}_{.\;m}.$ is accessed.
    \item Predictions are performed using a $\mathcal{M}_{m}$  model on $E_T$
  \end{enumerate}
\end{enumerate}
\item The step (\ref{CV1}) is repeated  until predictions are performed for all observations, and a matrix $\mathbf{M}_{VI}$ of importance is recorded.
\item The vector mean  $\widehat{VI}$ (representative)  of $\mathbf{M}_{VI}$ is determined.
\item The selection of each variable is done by $\widehat{VI}$ using  $VI_{min}$. 
\end{enumerate}
\label{LOLO_DCV_Algorithme}
\end{algorithm}

 \subsection{Variables selection strategies}
Four strategies of variables selection are implemented and compared to the reference method GLM-Lasso developped in \cite{Kouwayejds,Kouwayemldm}.
The first strategy, LDRT is a combination of LOLO-DCV (Leave-one-out-double-cross-validation), and regression tree (RT) \cite{Cart_Gey_Nedelec,Package_tree};
The second LDCT is a combination of LOLO-DCV and  conditional tree (CT) \cite{party}; The third 
LDRF, a combination of LOLO-DCV and Random forest (RF) \cite{Archer:2008:ECR:1316079.1316183, Random_forests_Biau_2010,Package_randomForest};  And the last LDCF is a combination of  LOLO-DCV and  conditional forest (CF) \cite{party}.
 The  selection power, the selection accuracy of each strategy is determined 
 on simulated data. 
The model construction parameter for each strategy is accessed based on the quadratic distance 
between importance and the mean of importance.
Each strategy based on the threshold of the minimum of variable importance selects an optimal subset of variables.

%
\section {Results} 
\label{Resultats}
\begin{figure}[!h]
\begin{center}
\includegraphics[width=4in,height=3.4in]{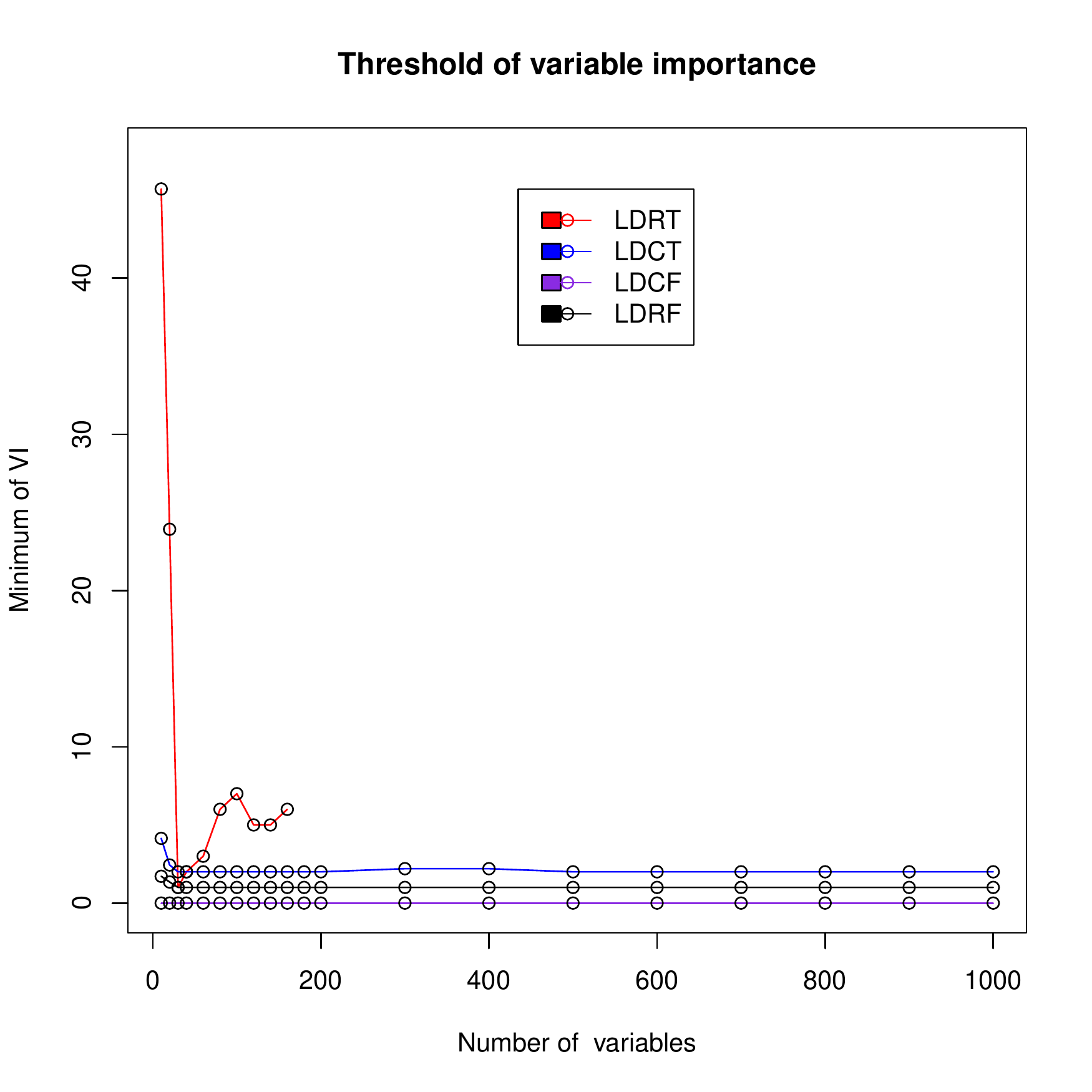}
\end{center}
\caption{\textbf{
Threshold of variable importance measurement. }Each line shows the trajectory of the 
variable importance for each strategy on simulated data.}
\label{Figure_Tree_Forest_Varmin}
\end{figure}

First of all, we show the numerical convergence of $VI_{min}$ in Equation \eqref{var_import_meas} for each strategy on 
simulated data. The results are shown in table \ref{Table_of_Selection_TF_Results} and  figure \ref{Figure_Tree_Forest_Varmin}. We also show the power of each strategy to reduce
 the number of variables, and  select  the true variables  in the optimal subset for prediction. The results are presented in table \ref{Table_of_Selection_TF_Results}
\subsection{Simulated study}
We simulated a data base of $n$ observations and $p$ true variables. We generated $p$-explanatory variables $X$, and the target variable
$Y$ knowing that $(Y|X) \sim \mathcal{P}(\mathbb{E} (X\beta))$ where $\beta$ is the vector of coefficients of $X$, and $\mathcal{P}(\mathbb{E} (X\beta))$ is a Poisson 
distribution of parameter $\mathbb{E} (X\beta)$.
We also generated another $p$-variables $Z$ which don't participate to the determination of $Y$. One of the strength of the algorithm is its capacity
to avoid in selection the wrong variables at most possible. The final number of variable in learning is $2\times p$
the set $X$ of generic explanatory variables contains : Gaussian variable, $X^p_{\mathcal{N}}\sim \mathcal{N}(\mu_p,\sigma_p^2)$;
discrete  variable which values are in range (1, 10), categorical variables with at most 10 modalities, and variables following Poisson distribution of parameter $\lambda_{X_\mathcal{P}}$.
For illustration, $\beta \sim \mathcal{N}(0,1)$, $\mu_p \in \{-1,0,1\},\sigma_p=1$, and $\lambda_{X_\mathcal{P}}=1$

\begin{figure}[!h]
\begin{center}
\includegraphics[width=4in,height=3.4in]{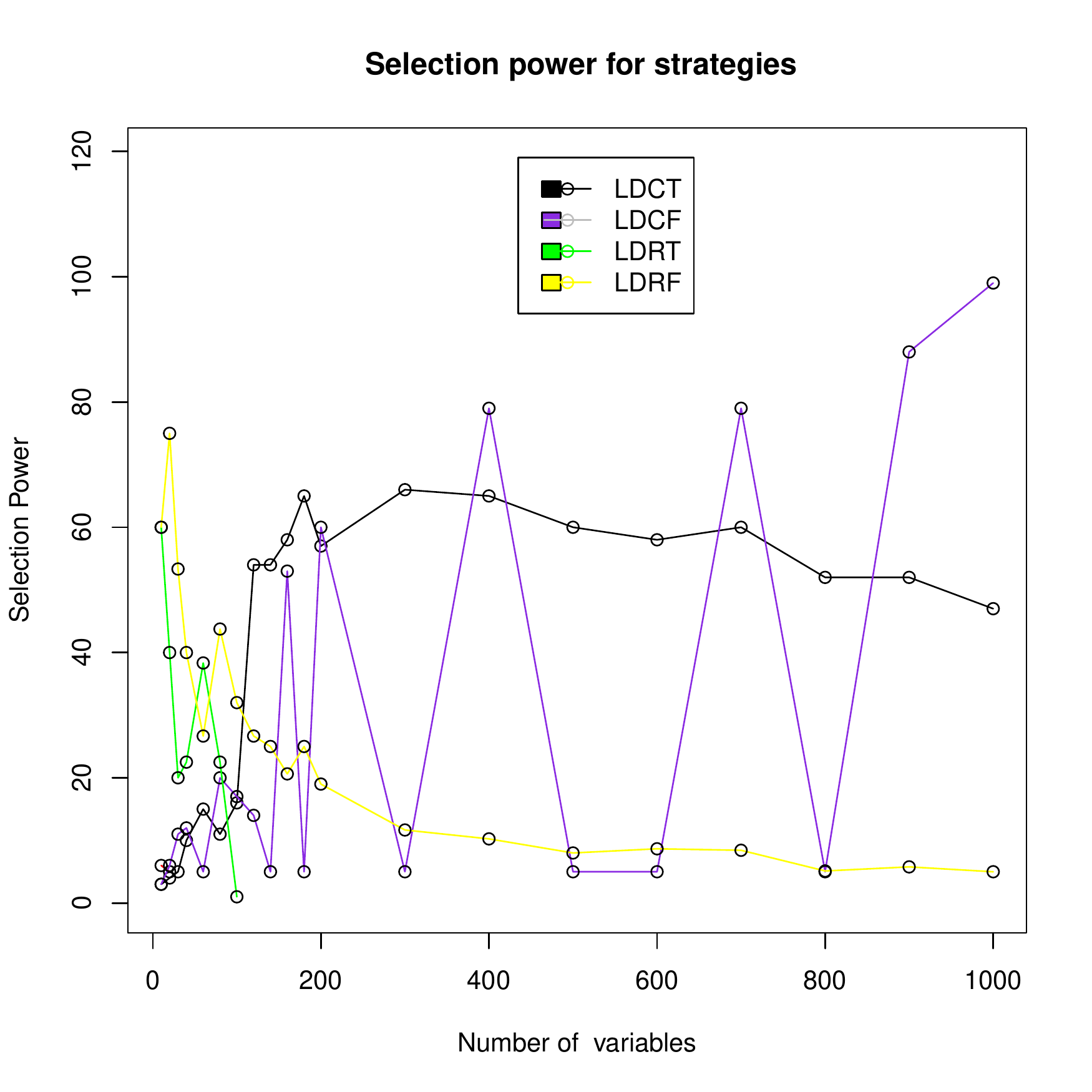}
\end{center}
\caption{\textbf{Selection power of strategies according  to number of variables.} Each line shows the trajectory of the 
selection for each strategy on simulated data.}
\label{Figure_Tree_Forest}
\end{figure}

\begin{table}[!ht]
\caption{\textbf{Summary on results of selection power and selection accuracy for 
different strategies.} MVI= Minimum variable importance, SP= selection power, SA= Selection accuracy} 

\begin{center}
\begin{tabular}{|l|r|r|r|r|r|r|r|r|r|}
\hline 
 number of variables& &250& 300 & 350 & 400 & 500 & 80 & 600 & 800 \cr
\hline 
LDRT&SP& - & - & - & - & - & - & - &- \cr
\cline{2-10}
&SA& - & - & - & - & - & - & - & - 
 \\
 \cline{2-10}
&MVI& - & - & - & - & - & - & - & - 
 \cr
 \hline 
LDCT&SP&  38.00 & 45 & 46 & 47.00 & 38.0 & 44.0 & 46.00 & 36.00 \cr
\cline{2-10}
&SA& 26.88 & 26.11 & 26.5 & 15.33 & 9.75 & 9 & 6.83 & 5.71 
 \\
 \cline{2-10}
&MVI&2.00 & 2 & 2.20 & 2.2 & 2.0 & 2.00 & 2.00 & 2 
 \cr
 \hline 
LDRF&SP& \textbf{20.62 }&  \textbf{21.67}&  \textbf{17.00} & \textbf{ 15.00} &  \textbf{11.75}
&  \textbf{9.00} &  \textbf{9.33} &  \textbf{8.57}  \cr
\cline{2-10}
&SA& 38.12 & 33.33 & 33 & 23 & 14.75 & 12 & 8.83 & 7.14 
 \\
 \cline{2-10}
&MVI&1.00 & 1.0 & 1 & 1.00 & 1 & 1.00 & 1.00 & 1.0
 \cr
 \hline 
LDCF&SP&  \textbf{48 }&  \textbf{62.00 }&  \textbf{5.0} &  \textbf{67.00} &  \textbf{88 }&  \textbf{5} &  \textbf{5.00} & \textbf{ 5.00 } \cr
\cline{2-10}
&SA& 30 & 34.44 & 2.5 & 22.33 & 22 & 1 & 0.83 & 0.71 
 \\
 \cline{2-10}
&MVI& 0 & 0 & 0 & 0 & 0 & 0 & 0 & 0 
 \cr
\hline 
\end{tabular} \vspace{0.2cm}\\
\label{Table_of_Selection_TF_Results} 
\end{center}
\end{table}

\subsection{Application to malaria data}
The results of  application of strategies LDRT, LDCT, LDRF, and LDCF on malaria data 
are shown in tables  \ref{var_impor_Moust}, and \ref{Table_LOLO_RF_malaria}.
 The strategy LDRT do not converge. The threshold of variable importance  is null  for LDCF. So any variable 
with  non null importance will be importante. The threshold of variable importance 
is very high for LDCT,  table \ref{var_impor_Moust}. The strategies LDCT, LDCF, and LDRF have a mean in prediction which is
equal to the mean of observations. LDRF has the low quadratic risk, abolute risk, the low computation time, and the most sparse 
subset of remained variables but its mean in prediction is greater than the one of observations, table \ref{Table_LOLO_RF_malaria}. 

%
%
%
%

\begin{table}
\caption{\textbf{Threshold of variable importance measurement on real data.}}

\begin{center}

\begin{tabular}{|l|r|r|r|r|}

\hline
&LDRT&LDCT&LDRF&LDCF\cr
\hline
  VI$_{min}$&-&5.12507&1.1&0\cr

\hline
\end{tabular} \vspace{0.2cm}\\
\label{var_impor_Moust} 
\end{center}
\end{table}

%
%
%
%
%

\begin{table}[!ht]
\caption{\textbf{Summary on results of selection power, selection accuracy, and minimum of variable importance on malaria data }}
\begin{center}
\begin{tabular}{|l|r|r|r|r|r|c|}
\hline
 Method & Mean & QR &Absolute risk& Remain variables  & Time CPU\\
 \hline
 Observations&3.74&-&-&-&-\cr
  \hline
  GLM-Lasso&3.74&54.54&3.669&3&25786.87\cr
  \hline
LDRT&-&-&-&-&-\cr
\hline
LDCT&3.74&50.56&3.217&4&24378.38\cr
\hline
LDRF& 3.75 &\textbf { 49.56}&\textbf {2.876} &\textbf {3} & \textbf {6715.09}\cr
\hline
LDCF&3.74&49.987&3.001&4&24830.34\cr
\hline
\end{tabular} \vspace{0.2cm}\\
\label{Table_LOLO_RF_malaria} 
\end{center}
\end{table}
\section{Discussion}
\label{Discussion}
The table \ref{Table_of_Selection_TF_Results} shows that computation is not compiled for LDRT if the number of variables is greater than 40 approximatively. 
This is due to the non convergence of the \textbf{rpart} function
in the package \textbf{rpart} for construction of regression tree. For Random forest, the convergence 
of the percentage of selected variables is not ensured. But the convergence is obtained when we combined  LOLO-DCV  with Random Forest (LDRF).
 The percentage of remained variables is around 5\%. The results are also shown in figure \ref{Figure_Tree_Forest}. It is evident that the 
 convergence of LDCF is not stable because the  alternative high and low percentage of remained variables.
 The figure \ref{Figure_Tree_Forest_Varmin} confirms the non convergence of LDRT over  approximatively 40 variables. Even if the algorithm
 did not converge for all number of variable, the minimum is attempted, and the trajectory is convex.
 For the strategies LDCT, LDRF, and  LDCF, the  minimum of variable importance converges, and  the results are shown in the line "MVI" of table
  \ref{Table_of_Selection_TF_Results}, and in figure \ref{Figure_Tree_Forest_Varmin} for each strategy. This denotes that the algorithm can compute correctly the threshold of importance 
  of variable for any number of variable but not for LDRT. 
  The results of application about malaria data  are shown in table \ref{Table_LOLO_RF_malaria}. The minimum of variable importance that we got for each 
  method are noted in table \ref{var_impor_Moust}. Unfortunately this minimum for LDCF is null. It means that for any positive value, the variable is important
  but four variable are selected at the end.
  For LDCT, the threshold of importance is very high, table \ref{var_impor_Moust} nevertheless four variables are remained  in the final model, tabe \ref{Table_LOLO_RF_malaria}.
  It  denotes that only few variables 
  are important in the model.
  The methods which  have  the mean  in prediction equals to the mean of observations  are LDCT, and  LDCF.  LDCF is  the best in  selection accuracy. 
  LDRF   has the  lowest quadratic risk, the lower absolute risk but it isn't the best   prediction. LDRF is the most sparse method with three variables. 
 Unlike LDRF,  LDCF, and  LDCT  which  are  more time consuming table \ref{Table_LOLO_RF_malaria}. 
 
\section{Conclusion}
In this work, we implemented an algorithm for the prediction of malaria risk  using 
environmental and climate variables. We performed the variables selection using an
automatic machine learning
 by a method combining regression trees or random forest,  and stratified two levels cross validation.
 The minimum threshold of variable importance is computed. variables selected by each 
 strategy are used to perform prediction. The results obtained with this method is clearly improved by those  
  obtained with the combination of Lasso, and LOLO-DCV  (GLM-Lasso) taken as  reference method. The improvement concerned 
  all properties such as the quality of the selection, and prediction. Moreover,
  this method didn't need  interaction between variables, 
 the pre-treatments 
 of experts were overcome, and the CPU time used to display
our program is smaller than the one 
 required by the reference method. The optimal subset of 
 variables for prediction contained season, mean rain fall, and vegetation index.
\section{Apendix}
Table of  variables Description
  \begin{table}[!h]
\caption{
\textbf{Description of variables}. Variables with star are recoded. \label{tab:Recoded variables}}
\begin{center}
\begin{tabular}{|l|l|c|l|} 
\hline
\textbf{} &\textbf{Nature}&\textbf{Number of modalities}&\textbf{Modalities}\cr
\hline
Repellent&Non-numeric& 2 & Yes/ No\cr 
\hline
Bed-net&Non-numeric &2 & Yes/  No\cr
\hline
Type of roof&Non-numeric& 2 & Sheet metal/ Straw\cr
\hline
Utensils& Non-numeric& 2 & Yes/  No\cr
\hline
Presence of constructions &Non-numeric& 2 & Yes/  No\cr
\hline
Type of soil &Non-numeric&2&Humid/ Dry\cr
\hline
Water course&Non-numeric&2&Yes/ No\cr
\hline
Majority class $^*$&Non-numeric&3&1/2/3\cr
\hline
Season &Non-numeric&4&1/2/3/4\cr
\hline
Village$^*$&Non-numeric&9&\cr
\hline
House $^*$&Non-numeric&41&\cr
\hline
Rainy days before  mission $^*$&Non-numeric&3&Quartile\cr
\hline
Rainy days during  mission &Numeric&Discrete&0/1/$\cdots$/3\cr
\hline
Fragmentation index $^*$&Non-numeric&4&Quartile\cr
\hline
Openings$^*$&Non-numeric&4&Quartile\cr
\hline
Nber of inhabitants $^*$&Non-numeric&3&Quartile\cr
\hline
Mean rainfall $^*$ &Non-numeric&4& Quartile\cr
\hline
Vegetation$^*$&Non-numeric&4&Quartile\cr
\hline
Total Mosquitoes &Numeric&Discrete&0/$\cdots$/481\cr
\hline
Total Anopheles&Numeric&Discrete&0/$\cdots$/87\cr
\hline
Anopheles infected &Numeric&Discrete&0/$\cdots$/9\cr
\hline
\end{tabular}
\end{center}
\end{table}
\bibliographystyle{elsarticle-num}
The authors have declared that no competing interests exist.
\newpage
\bibliography{Bibliography-Random_forest.bib}

\end{document}